\documentclass[10pt,twocolumn,letterpaper]{article}

\usepackage{amsmath}
\usepackage{amssymb}
\usepackage{caption}
\usepackage{comment}
\usepackage{cvpr}
\usepackage{epsfig}
\usepackage{graphicx}
\usepackage{subcaption}
\usepackage{times}
\usepackage[pagebackref=true,breaklinks=true,letterpaper=true,colorlinks,bookmarks=false]{hyperref}

\cvprfinalcopy % *** Uncomment this line for the final submission

\def\cvprPaperID{16} % *** Enter the CVPR Paper ID here

\newcommand{\TODO}[1]{{\color{red} #1}}

\ifcvprfinal\fi
\begin{document}

%%%%%%%%% TITLE
\title{Attention Mesh: High-fidelity Face Mesh Prediction in Real-time}

\author{
Ivan Grishchenko \quad Artsiom Ablavatski \quad Yury Kartynnik \quad Karthik Raveendran \quad Matthias Grundmann\\
Google Research\\
1600 Amphitheatre Pkwy, Mountain View, CA 94043, USA\\
{\tt\small \{igrishchenko, artsiom, kartynnik, krav, grundman\}@google.com}\\
}

\maketitle
%\thispagestyle{empty}

%%%%%%%%% ABSTRACT
\begin{abstract}

We present Attention Mesh, a lightweight architecture for 3D face mesh prediction that uses attention to semantically meaningful regions. Our neural network is designed for real-time on-device inference and runs at over 50 FPS on a Pixel 2 phone. Our solution enables applications like AR makeup, eye tracking and AR puppeteering that rely on highly accurate landmarks for eye and lips regions. Our main contribution is a unified network architecture that achieves the same accuracy on facial landmarks as a multi-stage cascaded approach, while being 30 percent faster.

\end{abstract}

%%%%%%%%% BODY TEXT
\section{Introduction}

In this work, we address the problem of registering a detailed 3D mesh template to a human face on an image. This registered mesh can be used for the virtual try-on of lipstick or puppeteering of virtual avatars where the accuracy of lip and eye contours is critical to realism.

In contrast to methods that use a parametric model of the human face \cite{Blanz3DMM}, we directly predict the positions of face mesh vertices in 3D. We base our architecture on earlier efforts in this field  \cite{FaceMesh} that use a two stage architecture involving a face detector followed by a landmark regression network. However, using a single regression network for the entire face leads to degraded quality in regions that are perceptually more significant (\eg lips, eyes).

One possible way to alleviate this issue is a cascaded approach: use the initial mesh prediction to produce tight crops around these regions and pass them to specialized networks to produce higher quality landmarks. While this directly addresses the problem of accuracy, it introduces performance issues, \eg relatively large separate models that use the original image as input, and additional synchronization steps between the GPU and CPU that are very costly on mobile phones. In this paper, we show that it is possible for a single model to achieve the same quality as the cascaded approach by employing region-specific heads that transform the feature maps with spatial transformers \cite{jaderberg2015spatial}, while being up to 30 percent faster during inference. We term this architecture as \emph{attention mesh}. An added benefit is that it is easier to train and distribute since it is internally consistent compared to multiple disparate networks that are chained together.

\begin{figure}[t]
    \centering
    \begin{subfigure}[b]{0.8\columnwidth}
        \frame{\includegraphics[width=\linewidth]{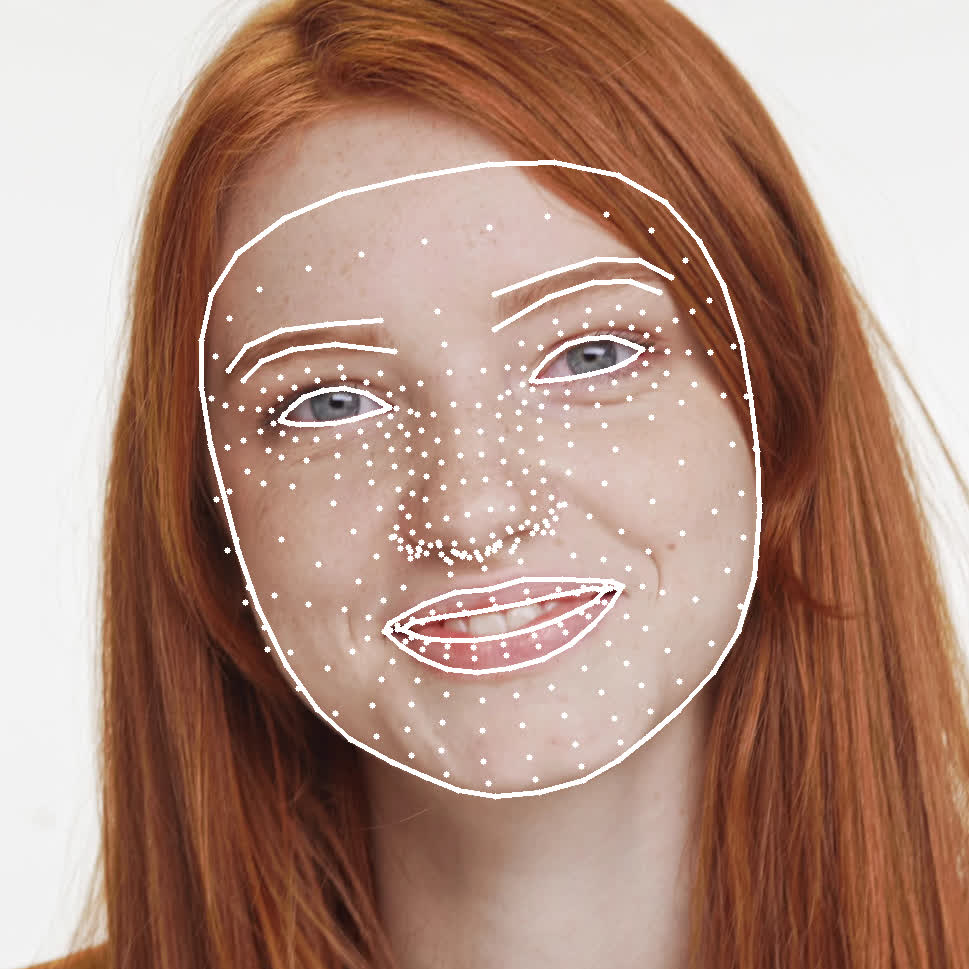}}
    \end{subfigure}
    \caption{\protect\centering Salient contours predicted by Attention Mesh submodels}
    \label{fig:salient_contours}
\end{figure}

\begin{comment}
\begin{figure}[t]
    \begin{center}
        \fbox{\rule{0pt}{2in} \rule{0.9\linewidth}{0pt}}
        %\includegraphics[width=0.8\linewidth]{egfigure.eps}
    \end{center}
    \caption{Attention Mesh prediction (Face with Mesh + zoomed lips and eyes with perfect quality)}
    \label{fig:long}
\end{figure}
\end{comment}

We use an architecture similar to one described in~\cite{DeepRegressionArchitecture}, where the authors build a network that is robust to the initialization provided by different face detectors. Despite the differing goals of the two papers, it is interesting to note that both suggest that a combination of using spatial transformers with heads corresponding to salient face regions produces marked improvements over a single large network. We provide the details of our implementation for producing landmarks corresponding to eyes, irises, and lips, as well as quality and inference performance benchmarks.

\begin{figure*}
\begin{center}
\end{center}
   \includegraphics[width=\linewidth]{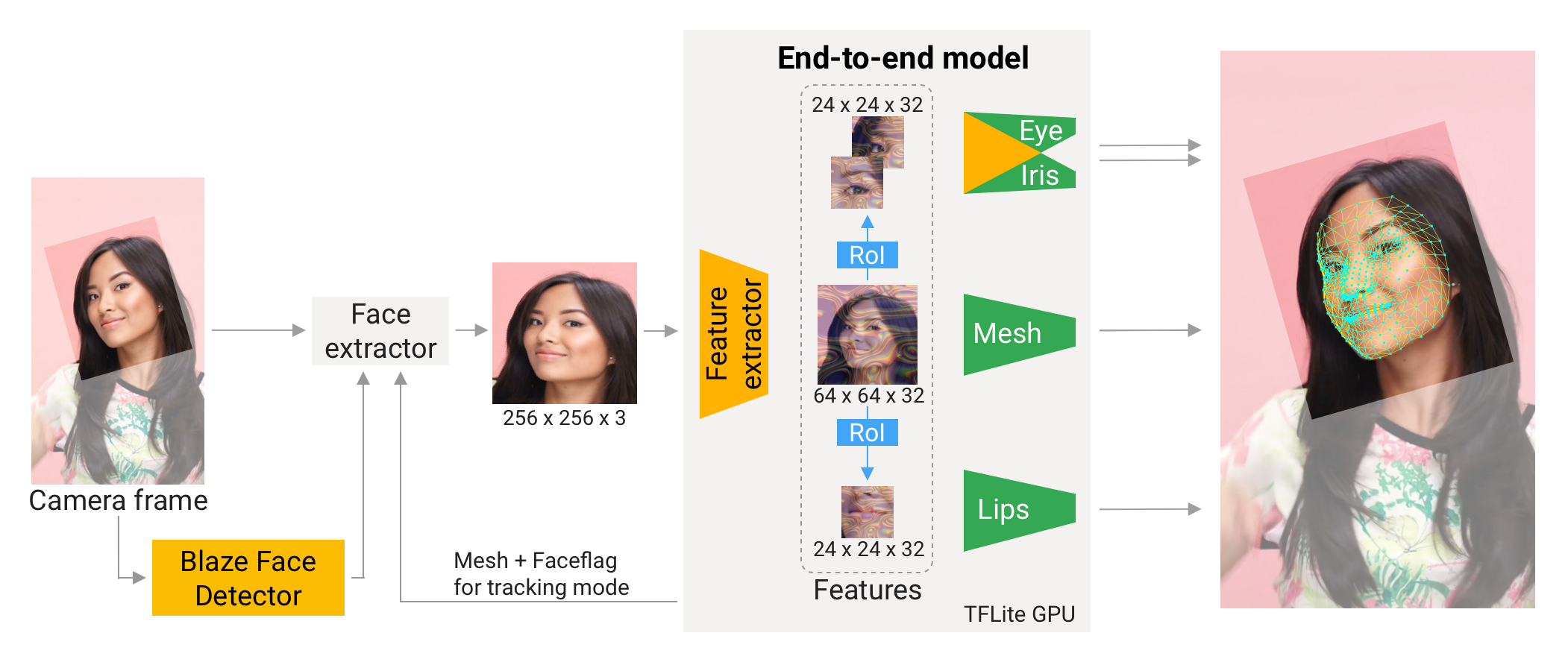}
   \caption{The inference pipeline and the model architecture overview}
   \label{fig:pipeline_n_architecture}
\label{fig:short}
\end{figure*}

%------------------------------------------------------------------------
\section{Attention mesh}

\paragraph{Model architecture}
The model accepts a $256 \times 256$ image as input. This image is provided by either the face detector or via tracking from a previous frame. After extracting a $64 \times 64$ feature map, the model  splits into several sub-models (Figure \ref{fig:pipeline_n_architecture}). One submodel predicts all 478 face mesh landmarks in 3D and defines crop bounds for each region of interest. The remaining submodels predict region landmarks from the corresponding $24\times24$ feature maps that are obtained via the attention mechanism.

We concentrate on three facial regions with key contours: the lips and two eyes (Figure \ref{fig:salient_contours}). Each eye submodel predicts the iris as a separate output after reaching the spatial resolution of $6\times6$. This allows the reuse of eye features while keeping dynamic iris independent from the more static eye landmarks.

Individual submodels allow us to control the network capacity dedicated to each region and boost quality where necessary. To further improve the accuracy of the predictions, we apply a set of normalizations to ensure that the eyes and lips are aligned with the horizontal and are of uniform size.

We train the attention mesh network in two phases. First, we employ ideal crops from the ground truth with slight augmentations and train all submodels independently. Then, we obtain crop locations from the model itself and train again to adapt the region submodels to them.

\begin{comment}
Each submodel (face, lips, and eyes) contains a number of bottlenecks similar to the recent work of Kartynnik~\etal~\cite{FaceMesh}.
\end{comment}

%-------------------------------------------------------------------------
\paragraph{Attention mechanism}
Several attention mechanisms (soft and hard) have been developed for visual feature extraction~\cite{gregor2015draw,jaderberg2015spatial}. These attention mechanisms sample a grid of 2D points in feature space and extract the features under the sampled points in a differentiable manner (\eg using 2D Gaussian kernels or affine transformations and differentiable interpolations). This allows to train architectures end-to-end and enrich the features that are used by the attention mechanism. Specifically, we use a spatial transformer module~\cite{jaderberg2015spatial} to extract $24\times24$ region features from the $64\times64$ feature map. The spatial transformer is controlled by an affine transformation matrix $\theta$ (Equation \ref{equation_theta}) and allows us to zoom, rotate, translate, and skew the sampled grid of points.

\begin{equation} \label{equation_theta}
  \theta = \begin{bmatrix} \;
 			x_{x} & sh_{x} & t_{x} \; \\
 			sh_{y} & s_{y} & t_{y} \; \\
  			\end{bmatrix} \;
\end{equation}

This affine transformation can be constructed either via supervised prediction of matrix parameters, or by computing them from the output of the face mesh submodel. 
\begin{figure}[h]
    \centering
    \includegraphics[width=0.9\linewidth]{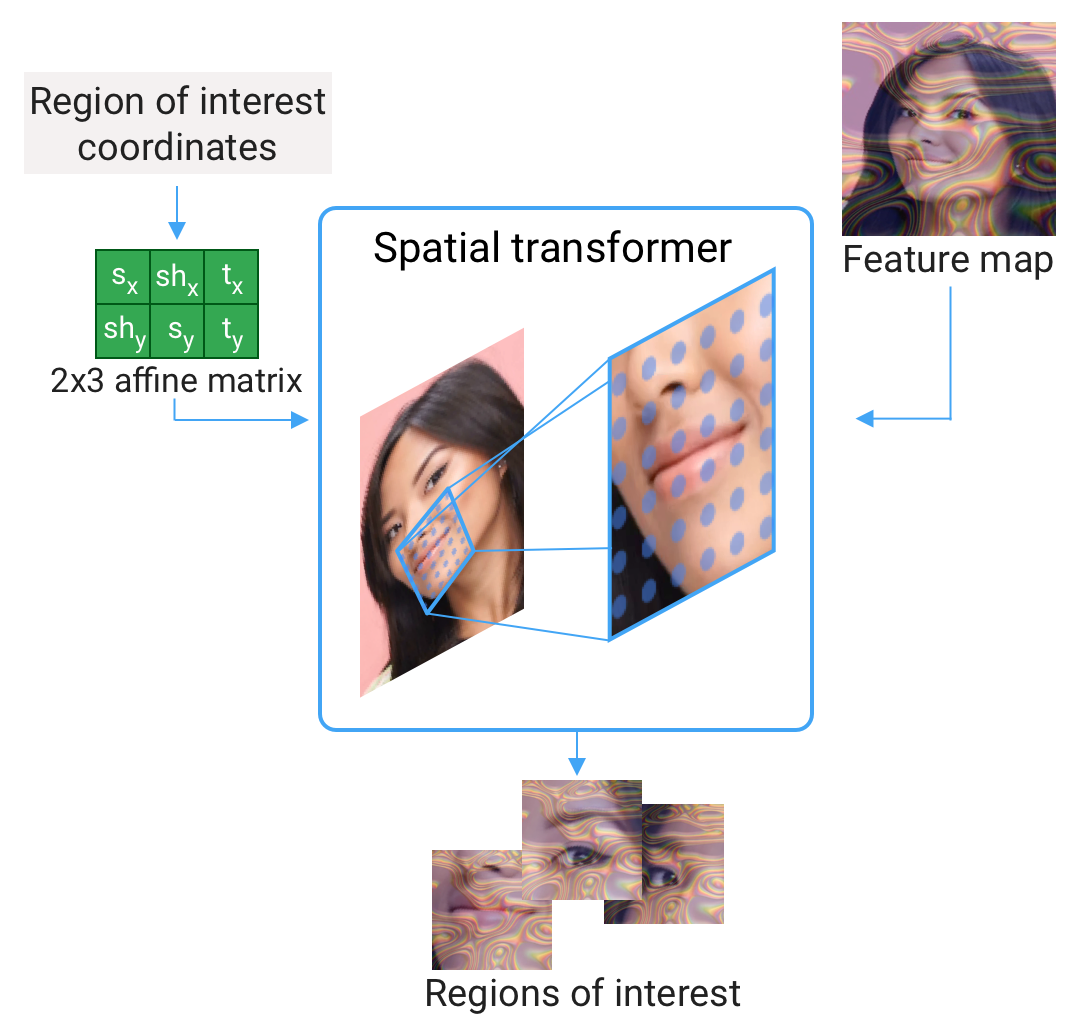}
    \caption{\protect\centering Spatial transformer as the attention mechanism}
    \label{fig:spatial_transformer}
\end{figure}

\paragraph{Dataset}

\begin{comment}
For the training, we used a dataset similar to the one introduced by Kartynnik~\etal\cite{FaceMesh}.
\end{comment}

Our dataset contains 30K in-the-wild mobile camera photos taken with numerous camera sensors and in varied conditions. We used manual annotation with special emphasis on consistency for salient contours to obtain the ground truth mesh vertex coordinates in 2D. The $Z$ coordinate was approximated using a synthetic model.

%------------------------------------------------------------------------
\section{Results}

To evaluate our unified approach, we compare it against the cascaded model which consists of independently trained region-specific models for the base mesh, eyes and lips that are run in succession. 

\iffalse
A standalone face mesh model for use in the cascaded approach can be easily extracted from the attention mesh without structure or weight modifications. Separate region models in the cascade have the same architecture as the face mesh submodel: namely, the same receptive field and the same architecture as the submodels in the attention mesh have after the downsampling to the $64\times64$ resolution.
\fi

%-------------------------------------------------------------------------
\paragraph{Performance}

Table~\ref{tbl:performance_eval} demonstrates that the attention mesh runs $25\%+$ faster than the cascade of separate face and region models on a typical modern mobile device. The performance has been measured using the TFLite GPU inference engine~\cite{lee2019device}. An additional $5\%$ speed-up is achieved due to the reduction of costly CPU-GPU synchronizations, since the whole attention mesh inference is performed in one pass on the GPU.

\begin{table}[ht]
\centering
\begin{tabular}{|l|c|c|c|c|}
\hline
Model &  Inference Time (ms) \\
\hline
Mesh & 8.82 \\
Lips & 4.18 \\
Eye \& iris & 4.70 \\
\hline
Cascade (sum of above) & 22.4 \\
\textbf{Attention Mesh} & \textbf{16.6}\\
\hline
\end{tabular}
\vskip 1ex
\caption{Performance on Pixel 2XL (ms)}
\label{tbl:performance_eval}
\end{table}

%-------------------------------------------------------------------------
\paragraph{Mesh quality}

A quantitative comparison of both models is presented in Table~\ref{tbl:quality_eval}. As the representative metric, we employ the mean distance between the predicted and ground truth locations of a specific subset of the points, normalized by 3D interocular distance (or the distance between the corners in the case of lips and eyes) for scale invariance. The attention mesh model outperforms the cascade of models on the eye regions and demonstrates comparable performance on the lips region. 

\iffalse
This change may be justified in practice as the face mesh submodel is not required to produce very accurate predictions for the lips and eye regions.
\fi

\begin{table}[ht]
\centering
\begin{tabular}{|l|c|c|c|c|}
\hline
Model &  All & Lips & Eyes \\
\hline
Mesh & 2.99 & 3.28 & 6.6 \\ \hline
Cascade & 2.99 & 2.70 & 6.28 \\ \hline
\textbf{Attention mesh} & \textbf{3.11} & \textbf{2.89} & \textbf{6.04} \\ \hline
\end{tabular}
\vskip 1ex
\caption{Mean normalized error in 2D.}
\label{tbl:quality_eval}
\end{table}

%------------------------------------------------------------------------
\section{Applications}
The performance of our model enables several real-time AR applications like virtual try-on of makeup and puppeteering. 

\begin{figure}[h]
    \centering
    \iffalse
    \begin{subfigure}[b]{.49\columnwidth}
        \frame{\includegraphics[width=\linewidth]{Makeup_1.png}}
    \end{subfigure}
    \fi
    \begin{subfigure}[b]{0.49\columnwidth}
        \frame{\includegraphics[width=\linewidth]{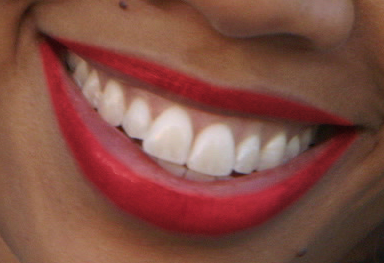}}
    \end{subfigure}
    \begin{subfigure}[b]{0.49\columnwidth}
        \frame{\includegraphics[width=\linewidth]{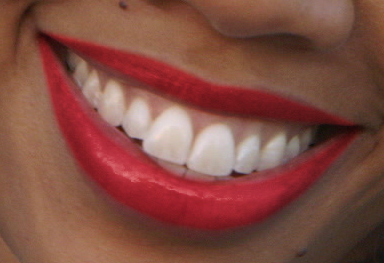}}
    \end{subfigure}
    \caption{\protect\centering Virtual makeup comparison: base mesh without refinements (left) vs. attention mesh with submodels (right)}
    \label{fig:makeup_application}
\end{figure}

\paragraph{AR Makeup}
Accurate registration of the face mesh is critical to applications like AR makeup where even small errors in alignment can drive the rendered effect into the "uncanny valley" \cite{TheUncannyValley}. We built a lipstick rendering solution (Figure \ref{fig:makeup_application}) on top of our attention mesh model by using the contours provided by the lip submodel. A/B testing on 10 images and 80 people showed that 46\% of AR samples were actually classified as real and 38\% of real samples --- as AR.%
\begin{comment}%
\footnote{Detailed analysis in \cite{ARLipstick}}
\end{comment}

\begin{figure}[h]
    \centering
    \begin{subfigure}[b]{0.49\columnwidth}
        \frame{\includegraphics[width=\linewidth]{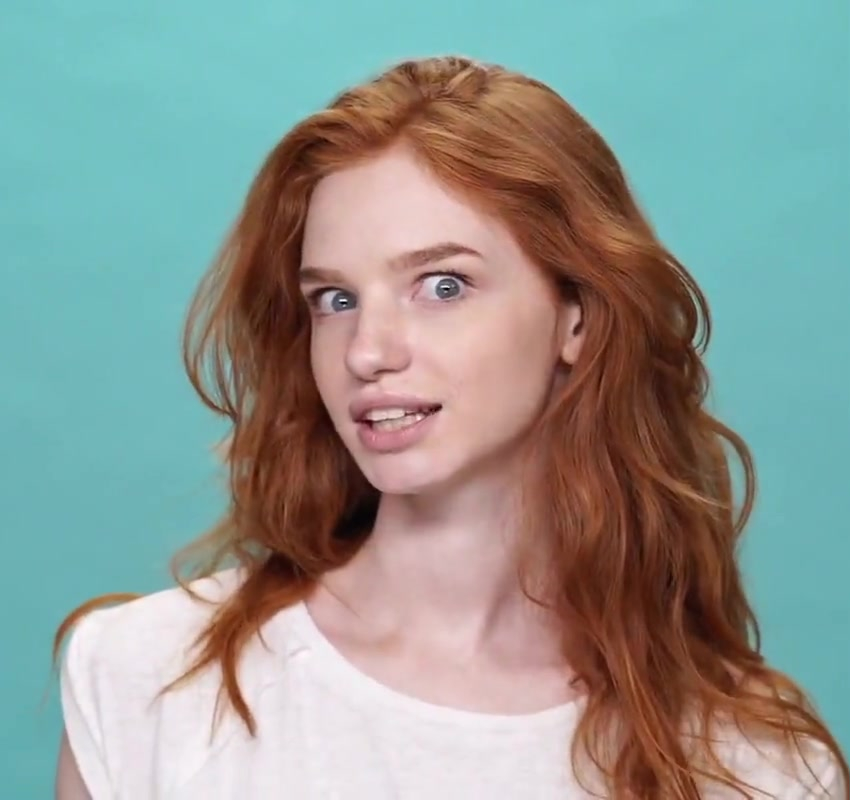}}
    \end{subfigure}
    \begin{subfigure}[b]{0.49\columnwidth}
        \frame{\includegraphics[width=\linewidth]{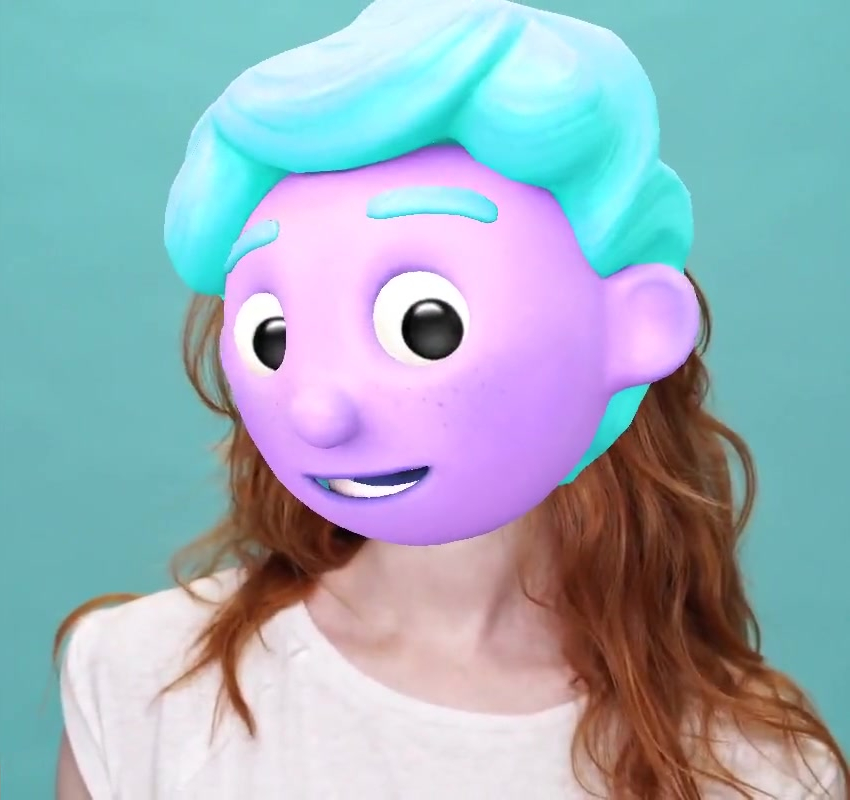}}
    \end{subfigure}
    \caption{\protect\centering Puppeteering}
    \label{fig:puppeteering_application}
\end{figure}

%------------------------------------------------------------------------
\paragraph{Puppeteering}
Our model can also be used for virtual puppeteering and facial triggers. We built a small fully connected model that predicts 10 blend shape coefficients for the mouth and 8 blend shape coefficients for each eye. We feed the output of the attention mesh submodels to this blend shape network. In order to handle differences between various human faces, we apply Laplacian mesh editing to morph a canonical mesh into the predicted mesh \cite{DualLaplacianMorphing}. This lets us use the blend shape coefficients for different human faces without additional fine-tuning. We demonstrate some results in Figure \ref{fig:puppeteering_application}.

\section{Conclusion}
We present a unified model that enables accurate face mesh prediction in real-time. By using a differentiable attention mechanism, we are able to devote computational resources to salient face regions without incurring the performance penalty of running independent region-specific models.
Our model and demos will soon be available in MediaPipe (\url{https://github.com/google/mediapipe}).

%------------------------------------------------------------------------
\iffalse
\paragraph{Eye tracking}
Since Attention Mesh has enabled prediction of irises, it gives the opportunity to implement eye tracking and gaze detection. Both have a wide list of applications, including remote control in accessibility scenarios, car driving safety detection, market and UX research, gaming.
\TODO{Image of in-car gaze detection? Or just nothing?}
\fi

{\small
\bibliographystyle{ieee_fullname}
\bibliography{egbib}
}

\end{document}